# Compliance error compensation in robotic-based milling

Alexandr Klimchik[1,2], Dmitry Bondarenko[1,2], Anatol Pashkevich[1,2], Sébastien Briot[2], Benoît Furet[2,3]

[1]Ecole des Mines de Nantes, 4 rue Alfred-Kastler, 44307 Nantes, France
[2]Institut de Recherches en Communications et Cybernétique de Nantes, France
[3]Université de Nantes, quai de Tourville, 44035 Nantes, France
```
{alexandr.klimchik, anatol.pashkevich}@mines-nantes.fr,
{dmitry.bondarenko, sebastien.briot, benoit.furet}@irccyn.ec-
                            nantes.fr
```

**Abstract.** The paper deals with the problem of compliance errors compensation in robotic-based milling. Contrary to previous works that assume that the forces/torques generated by the manufacturing process are constant, the interaction between the milling tool and the workpiece is modeled in details. It takes into account the tool geometry, the number of teeth, the feed rate, the spindle rotation speed and the properties of the material to be processed. Due to high level of the disturbing forces/torques, the developed compensation technique is based on the non-linear stiffness model that allows us to modify the target trajectory taking into account nonlinearities and to avoid the chattering effect. Illustrative example is presented that deals with robotic-based milling of aluminum alloy.

## 1 Introduction

Currently, robots become more and more popular for a variety of technological processes, including high-speed precision machining. For this process, external loading caused by the machining force is applied on the robot tool. This force is generated by the interaction between the tool mounted on the robot end-effector and the workpiece during the material removal [1]. It is a contact force and it is distributed along the affected area of the tool cutting part. To evaluate the influence and to analyze the robot behavior while machining, the cutting force should be defined either experimentally or using accurate mathematical model.

To evaluate the force caused by interaction between the tool and the workpiece, two approaches can be used. The static approach allows computing the average cutting force without any consideration of dynamic aspect in machining system. This force serves as an external loading of the robot. This approach is widely used in analysis of conventional machining processes using CNC machines [2], where the stiffness is high. In contrast, robots have relatively low structural stiffness. For this rea-son, in the case of robotic-based machining, an additional source of dynamic displacements of the end-effector with respect to the desired trajectory induced by robot compliance may arise. Such behavior leads to the variable contact between the machining tool and the workpiece. Thus, the generated contact force depends on the current position of the robot end-effector on the trajectory. Consequently, the cutting force cannot be evaluated correctly using the static approach. In this case, the dynamic approach, which will be used in the paper, is required. It is based on computing of the force at each instant of machining process that defines loading of the robot for the next instant of processing. As a result, the dynamic aspect of robot motion under such variable cutting force can be examined for whole process.

Usually, in the robot-based machining this force causes essential deflections that decrease the quality of the final product. The problem of the robot error compensation can be solved in two ways that differ in degree of modification of the robot control software:

(a) by *modification of the manipulator model*, which better suits to the real manipulator and is used by the robot controller (in simple case, it can be limited by tuning of the nominal manipulator model, but may also involve essential model enhancement by introducing additional parameters, if it is allowed by a robot manufacturer);

(b) by *modification of the robot control program* that defines the prescribed trajectory in Cartesian space (here, using relevant error model, the input trajectory is generated in such way that under the loading the output trajectory coincides with the desired one, while input trajectory differs from the target one).

Moreover, with regard to the robot-based machining, there is a solution that does not require force/torque measurements or computations [1], where the target trajectory for the robot controller is modified by applying the "mirror" technique. An evident advantage of this technique is its applicability to the compensation of all types of the robot errors, including geometrical and compliance ones. However, this approach requires carrying out additional preliminary experiments which are quite expensive. So, it is suitable for the large-scale production only. Another compensation methodology has been proposed by Eastwood and Webb [3] that was used for gravitational deflection compensation for hybrid parallel kinematic machines.

This paper focuses on the modification of control program that is considered to be more realistic in practice. This approach requires also accurate stiffness model of the manipulator. From point of view of stiffness analysis, the external and forces directly influence on the manipulator equilibrium configuration and, accordingly, may modify the stiffness properties. So, they must be undoubtedly taken into account while developing the stiffness model. However, in most of the related works the Cartesian stiffness matrix has been computed for the nominal configuration [4-5]. Such approach is suitable for the case of small deflections only. For the opposite case, the most important results have been obtained in [6-8], which deal with the stiffness analysis of manipulators under the end-point loading.

Thus, to compensate errors caused by the machining process, it is required to have an accurate stiffness model and precise cutting force model. In contrast to the previous works, the compliance error compensation technique presented in this work is based on the non-linear stiffness model of the manipulator [7] and dynamic model of technological process that generates the cutting force.

## 2   Problem statement

For the *compliance errors*, the compensation technique must rely on two components. The first of them describes distribution of the stiffness properties throughout the workspace and is defined by the stiffness matrix as a function of the joint coordinates. The second component describes the forces/torques acting on the end-effector while the manipulator is performing its machining task (manipulator *loading*).

The *stiffness matrix* required for the compliance errors compensation highly depends on the robot configuration and essentially varies throughout the workspace. From general point of view, full-scale compensation of the compliance errors requires essential revision of the manipulator model embedded in the robot controller. In fact, instead of conventional geometrical model that provides inverse/direct coordinate transformations from the joint to Cartesian spaces and vice versa, here it is necessary to employ the so-called *kinetostatic model* [9]. It is essentially more complicated than the geometrical model and requires rather intensive computations that are presented in Section 3.

The dynamic behavior of the robot under the loading $\mathbf{F}$ caused by technological process can be described as

$$\mathbf{M}_C \ddot{\boldsymbol{\delta} \mathbf{t}} + \mathbf{C}_C \dot{\boldsymbol{\delta} \mathbf{t}} + \mathbf{K}_C \boldsymbol{\delta} \mathbf{t} = \mathbf{F} \qquad (1)$$

where $\mathbf{M}_C$ is $6 \times 6$ mass matrix that represents the global behavior of the robot in terms of natural frequencies, $\mathbf{C}_C$ is $6 \times 6$ damping matrix, $\mathbf{K}_C$ is $6 \times 6$ Cartesian stiffness matrix of the robot under the external loading $\mathbf{F}$, $\boldsymbol{\delta} \mathbf{t}, \dot{\boldsymbol{\delta} \mathbf{t}}$ and $\ddot{\boldsymbol{\delta} \mathbf{t}}$ are dynamic displacement, velocity and acceleration of the tool end-point in a current moment respectively [10].

In general, the cutting force $F_c$ has a *nonlinear* nature and depends on many factors such as cutting conditions, properties of workpiece material and tool cutting part, etc [11]. But, for given tool/workpiece combination, the force $F_c$ could be approximated *as a function* of an uncut chip thickness $h$, which represents the desired thickness to cut at each instant of machining.

Hence, to reduce the errors caused by the cutting forces in the robotic-based machining it is required to obtain an accurate elastostatic model of the robot and elasto-dynamic model of the machining process. These problems are addressed in the following sections taking into account some particularities of the considered application (robotic-based milling).

## 3 Manipulator model

### 3.1 Elastostatic model

Elastostatic model of a serial robot is usually defined by its Cartesian stiffness matrix, which should be computed in the neighborhood of loaded configuration. Let us propose numerical technique for computing static equilibrium configuration for a general type of serial manipulator. Such manipulator may be approximated as a set of rigid links and virtual joints, which take into account elastostatic properties (Fig. 1). Since the link weight of serial robots is not negligible, it is reasonable to decompose it into two parts (based on the link mass centre) and apply them to the both ends of the link. All this loadings will be aggregated in a vector $\mathbf{G} = [\mathbf{G}_1...\mathbf{G}_n]$, where $\mathbf{G}_i$ is the loading applied to the $i$-th node-point. Besides, it is assumed that the external loading $\mathbf{F}$ (caused by the interaction of the tool and the workpiece) is applied to the robot end-effector.

Following the principle of virtual work, the work of external forces $\mathbf{G}$, $\mathbf{F}$ is equal to the work of internal forces $\boldsymbol{\tau}_\theta$ caused by displacement of the virtual springs $\delta\boldsymbol{\theta}$

$$\sum_{j=1}^{n}\left(\mathbf{G}_j^\mathrm{T}\cdot\delta\mathbf{t}_j\right)+\mathbf{F}^\mathrm{T}\cdot\delta\mathbf{t}=\boldsymbol{\tau}_\theta^\mathrm{T}\cdot\delta\boldsymbol{\theta} \qquad (2)$$

where the virtual displacements $\delta\mathbf{t}_j$ can be computed from the linearized geometrical model derived from $\delta\mathbf{t}_j = \mathbf{J}_\theta^{(j)}\delta\boldsymbol{\theta}, j=1..n$, which includes the Jacobian matrices $\mathbf{J}_\theta^{(j)} = \partial\mathbf{g}_j(\mathbf{q},\boldsymbol{\theta})/\partial\boldsymbol{\theta}$ with respect to the virtual joint coordinates.

So, expression (2) can be rewritten as

$$\sum_{j=1}^{n}\left(\mathbf{G}_j^\mathrm{T}\cdot\mathbf{J}_\theta^{(j)}\cdot\delta\boldsymbol{\theta}\right)+\left(\mathbf{F}^\mathrm{T}\cdot\mathbf{J}_\theta^{(n)}\cdot\delta\boldsymbol{\theta}\right)=\boldsymbol{\tau}_\theta^\mathrm{T}\cdot\delta\boldsymbol{\theta} \qquad (3)$$

which has to be satisfied for any variation of $\delta\boldsymbol{\theta}$. It means that the terms regrouping the variables $\delta\boldsymbol{\theta}$ have the coefficients equal to zero. Hence the force balance equations can be written as

$$\boldsymbol{\tau}_\theta = \sum_{j=1}^{n}\mathbf{J}_\theta^{(j)\mathrm{T}}\cdot\mathbf{G}_j + \mathbf{J}_\theta^{(n)\mathrm{T}}\cdot\mathbf{F} \qquad (4)$$

These equations can be re-written in block-matrix form as

$$\tau_\theta = \mathbf{J}_\theta^{(G)T} \cdot \mathbf{G} + \mathbf{J}_\theta^{(F)T} \cdot \mathbf{F} \qquad (5)$$

where $\mathbf{J}_\theta^{(F)} = \mathbf{J}_\theta^{(n)}$, $\mathbf{J}_\theta^{(G)} = \left[ \mathbf{J}_\theta^{(1)T} ... \mathbf{J}_\theta^{(n)T} \right]^T$, $\mathbf{G} = \left[ \mathbf{G}_1^T ... \mathbf{G}_n^T \right]^T$. Finally, taking into account the virtual spring reaction $\tau_\theta = \mathbf{K}_\theta \cdot \mathbf{\theta}$, where $\mathbf{K}_\theta = diag\left(\mathbf{K}_{\theta_1}, ..., \mathbf{K}_{\theta_n}\right)$, the desired static equilibrium equations can be presented as

$$\mathbf{J}_\theta^{(G)T} \cdot \mathbf{G} + \mathbf{J}_\theta^{(F)T} \cdot \mathbf{F} = \mathbf{K}_\theta \cdot \mathbf{\theta} \qquad (6)$$

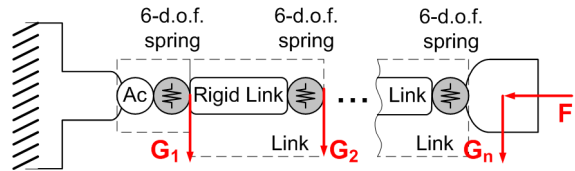

**Fig. 1.** VJM model of industrial robot with end-point and auxiliary loading

To obtain a relation between the external loading $\mathbf{F}$ and internal coordinates of the kinematic chain $\mathbf{\theta}$ corresponding to the static equilibrium, equations (6) should be solved either for different given values of $\mathbf{F}$ or for different given values of $\mathbf{t}$. Let us solve the static equilibrium equations with respect to the manipulator configuration $\mathbf{\theta}$ and the external loading $\mathbf{F}$ for given end-effector position $\mathbf{t} = \mathbf{g}(\mathbf{\theta})$ and the function of auxiliary-loadings $\mathbf{G}(\mathbf{\theta})$

$$\mathbf{K}_\theta \cdot \mathbf{\theta} = \mathbf{J}_\theta^{(G)T} \mathbf{G} + \mathbf{J}_\theta^{(F)T} \mathbf{F}; \quad \mathbf{t} = \mathbf{g}(\mathbf{\theta}); \quad \mathbf{G} = \mathbf{G}(\mathbf{\theta}) \qquad (7)$$

where the unknown variables are $(\mathbf{\theta}, \mathbf{F})$.

Since usually this system has no analytical solution, iterative numerical technique can be applied. So, the kinematic equations may be linearized in the neighborhood of the current configuration $\mathbf{\theta}_i$

$$\mathbf{t}_{i+1} = \mathbf{g}(\mathbf{\theta}_i) + \mathbf{J}_\theta^{(F)}(\mathbf{\theta}_i) \cdot (\mathbf{\theta}_{i+1} - \mathbf{\theta}_i); \qquad (8)$$

where the subscript '$i$' indicates the iteration number and the changes in Jacobians $\mathbf{J}_\theta^{(G)}$, $\mathbf{J}_\theta^{(F)}$ and the auxiliary loadings $\mathbf{G}$ are assumed to be negligible from iteration to iteration. Correspondingly, the static equilibrium equations in the neighborhood of $\mathbf{\theta}_i$ may be rewritten as

$$\mathbf{J}_\theta^{(G)T} \cdot \mathbf{G} + \mathbf{J}_\theta^{(F)T} \cdot \mathbf{F}_{i+1} = \mathbf{K}_\theta \cdot \mathbf{\theta}_{i+1} \qquad (9)$$

Thus, combining (8), (9) and expression for $\mathbf{\theta} = \mathbf{K}_\theta^{-1}(\mathbf{J}_\theta^{(G)T} \cdot \mathbf{G} + \mathbf{J}_\theta^{(F)T} \cdot \mathbf{F})$, the unknown variables $\mathbf{F}$ and $\mathbf{\theta}$ can be computed using following iterative scheme

$$\begin{aligned} \mathbf{F}_{i+1} &= \left(\mathbf{J}_\theta^{(F)} \cdot \mathbf{K}_\theta^{-1} \cdot \mathbf{J}_\theta^{(F)T}\right)^{-1} \left(\mathbf{t}_{i+1} - \mathbf{g}(\mathbf{\theta}_i) + \mathbf{J}_\theta^{(F)} \mathbf{\theta}_i - \mathbf{J}_\theta^{(F)} \mathbf{K}_\theta^{-1} \mathbf{J}_\theta^{(G)T} \mathbf{G}_i \right) \\ \mathbf{\theta}_{i+1} &= \mathbf{K}_\theta^{-1} \left(\mathbf{J}_\theta^{(G)T} \cdot \mathbf{G}_i + \mathbf{J}_\theta^{(F)T} \cdot \mathbf{F}_{i+1}\right) \end{aligned} \qquad (10)$$

The proposed algorithm allows us to compute the static equilibrium configuration for the serial robot under external loadings applied to any point of the manipulator and the loading from the technological process.

### 3.2 Stiffness matrix

In order to obtain the Cartesian stiffness matrix, let us linearize the force-deflection relation in the neighborhood of the equilibrium. Following this approach, two equili-briums that correspond to the manipulator state variables $(\mathbf{F}, \mathbf{\theta}, \mathbf{t})$ and $(\mathbf{F}+\delta\mathbf{F}, \mathbf{\theta}+\delta\mathbf{\theta}, \mathbf{t}+\delta\mathbf{t})$ should be considered simultaneously. Here, notations $\delta\mathbf{F}$, $\delta\mathbf{t}$ define small increments of the external loading and relevant displacement of the end-point. Finally, the static equilibrium equations may be written as

$$\mathbf{t} = \mathbf{g}(\mathbf{\theta}); \qquad \mathbf{K}_\theta \cdot \mathbf{\theta} = \mathbf{J}_\theta^{(G)T} \cdot \mathbf{G} + \mathbf{J}_\theta^{(F)T} \cdot \mathbf{F} \qquad (11)$$

and

$$\begin{aligned}
\mathbf{t} + \delta\mathbf{t} &= \mathbf{g}(\mathbf{\theta}+\delta\mathbf{\theta}) \\
\mathbf{K}_\theta \cdot (\mathbf{\theta}+\delta\mathbf{\theta}) &= \left(\mathbf{J}_\theta^{(G)} + \delta\mathbf{J}_\theta^{(G)}\right)^T \cdot (\mathbf{G}+\delta\mathbf{G}) + \left(\mathbf{J}_\theta^{(F)} + \delta\mathbf{J}_\theta^{(F)}\right)^T \cdot (\mathbf{F}+\delta\mathbf{F})
\end{aligned} \qquad (12)$$

where $\mathbf{t}$, $\mathbf{F}$, $\mathbf{G}$, $\mathbf{K}_\theta$, $\mathbf{\theta}$ are assumed to be known.

After linearization of the function $\mathbf{g}(\mathbf{\theta})$ in the neighborhood of the loaded equilibrium, the system (11), (12) is reduced to equations

$$\begin{aligned}
\delta\mathbf{t} &= \mathbf{J}_\theta^{(F)} \delta\mathbf{\theta} \\
\mathbf{K}_\theta \cdot \delta\mathbf{\theta} &= \delta\mathbf{J}_\theta^{(G)} \mathbf{G} + \mathbf{J}_\theta^{(G)} \delta\mathbf{G} + \delta\mathbf{J}_\theta^{(F)} \mathbf{F} + \mathbf{J}_\theta^{(F)} \delta\mathbf{F}
\end{aligned} \qquad (13)$$

which defines the desired linear relations between $\delta\mathbf{t}$ and $\delta\mathbf{F}$. In this system, small variations of Jacobians may be expressed via the second order derivatives $\delta\mathbf{J}_\theta^{(F)} = \mathbf{H}_{\theta\theta}^{(F)} \cdot \delta\mathbf{\theta}$, $\delta\mathbf{J}_\theta^{(G)} = \mathbf{H}_{\theta\theta}^{(G)} \cdot \delta\mathbf{\theta}$, where

$$\mathbf{H}_{\theta\theta}^{(G)} = \sum_{j=1}^{n} \partial^2 \mathbf{g}_j^T \mathbf{G}_j \big/ \partial\mathbf{\theta}^2; \qquad \mathbf{H}_{\theta\theta}^{(F)} = \partial^2 \mathbf{g}^T \mathbf{F} \big/ \partial\mathbf{\theta}^2 \qquad (14)$$

Also, the auxiliary loading $\mathbf{G}$ may be computed via the first order derivatives as $\delta\mathbf{G} = \partial\mathbf{G}/\partial\mathbf{\theta} \cdot \delta\mathbf{\theta}$

Further, let us introduce additional notation

$$\mathbf{H}_{\theta\theta} = \mathbf{H}_{\theta\theta}^{(F)} + \mathbf{H}_{\theta\theta}^{(G)} + \mathbf{J}_\theta^{(G)T} \cdot \partial\mathbf{G}/\partial\mathbf{\theta} \qquad (15)$$

which allows us to present system (13) in the form

$$\begin{bmatrix} \delta\mathbf{t} \\ \mathbf{0} \end{bmatrix} = \begin{bmatrix} \mathbf{0} & \mathbf{J}_\theta^{(F)} \\ \mathbf{J}_\theta^{(F)T} & -\mathbf{K}_\theta + \mathbf{H}_{\theta\theta} \end{bmatrix} \cdot \begin{bmatrix} \delta\mathbf{F} \\ \delta\mathbf{\theta} \end{bmatrix} \qquad (16)$$

So, the desired Cartesian stiffness matrices $\mathbf{K}_C$ can be computed as

$$\mathbf{K}_C = \left(\mathbf{J}_\theta^{(F)} (\mathbf{K}_\theta - \mathbf{H}_{\theta\theta})^{-1} \mathbf{J}_\theta^{(F)T}\right)^{-1} \qquad (17)$$

Below, this expression will be used for computing of the elastostatic deflections of the robotic manipulator.

### 3.3 Reduced mass matrix

To evaluate the dynamic behavior of the robot under the loading, in addition to the Cartesian stiffness matrix $\mathbf{K}_C$ it is required to define the Cartesian mass matrix $\mathbf{M}_C$. This mass matrix has the same dimension as $\mathbf{K}_C$ and can be obtained using some model reduction techniques. Comprehensive analysis and definition of this matrix have been proposed in [10]. Here, let us summarize the main results that will be used further.

To reduce the mass matrix dimension, model reduction techniques are applied for decreasing the size of the link mass matrices and also for the robot total mass matrix. Two main ways can be followed to reduce the size of the link mass matrices. The first one consists in discretizing the beam $j$ into $p_j$ rigid links and springs and to express their displacements as a function of the beam extremity displacements. However, such numerical method must be repeated for each link and, thus, increases the size of the algorithm and decreases its efficiency. As a result, it is preferred to use the following procedure which allows analytical expressions to be obtained for the reduced link mass matrices.

Let us consider the link $j$, modeled as a beam (Fig. 2). At this beam is attached a local frame represented by the vectors $\mathbf{x_j}$, $\mathbf{y_j}$ and $\mathbf{z_j}$. Before any deformation of the system, the beam $j$ is linked to beams ($j$–1) and ($j$+1) at points $O_j$ and $O_{j+1}$, respectively (Fig. 2). After deformation of the robot, the beam extremity located at $O_j$ is displaced from $\boldsymbol{\delta t}_{j-1} = \left[\delta t_{j-1}^1, \delta t_{j-1}^2, \ldots, \delta t_{j-1}^6\right]^T$ and the one located at $O_{j+1}$ is displaced from $\boldsymbol{\delta t}_j = \left[\delta t_j^1, \delta t_j^2, \ldots, \delta t_j^6\right]^T$, where the three first components of each vector correspond to the translational displacements along local $\mathbf{x_j}$, $\mathbf{y_j}$ and $\mathbf{z_j}$ axes, respectively, and the three last components to the rotational displacements along the same axes.

The general formula for the kinetic energy of an elastic Bernoulli beam is equal to:

$$T_j = 1/2 \int_0^{L_j} \rho_j \dot{\boldsymbol{\delta}}_j^T \mathbf{Q}_j \dot{\boldsymbol{\delta}}_j \, dx; \qquad \mathbf{Q}_j = \mathrm{diag}\left(A_j, A_j, A_j I_j^p, I_j^y, I_j^z\right) \qquad (18)$$

In this expression, $\dot{\boldsymbol{\delta}}_j$ represents the velocity of the beam cross-section located at position $x$ from the local reference frame (Fig. 2), $L_j$ is the length of the beam $j$, $\rho_j$ the mass density at cross-section $x$, $A_j$ its area, $I_j^p$ its torsional constant and $I_j^y$, $I_j^z$, the quadratic momentums along $\mathbf{y_j}$ and $\mathbf{z_j}$, respectively.

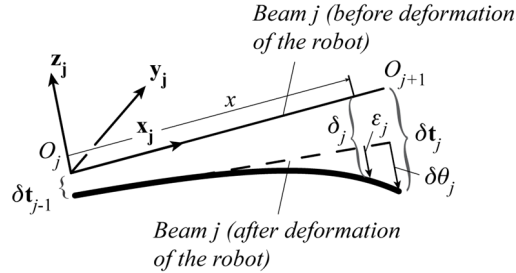

**Fig. 2.** Displacements and elastic deformations of a beam.

For the $l$-th natural mode $\omega_l$, the kinetic energy can be rewritten as:

$$T_{jl} = 1/2\, \omega_l^2 \cos^2(\omega_l t + \varphi_l) \int_0^{L_j} \rho_j \boldsymbol{\delta}_j^T \mathbf{Q}_j \boldsymbol{\delta}_j\, dx \tag{19}$$

$\boldsymbol{\delta}_j$ being the amplitude of the displacement of the beam cross-section located at position $x$ from the local reference frame (Fig. 2).

In the Rayleigh-Ritz approximation, considering that the deformations due to the natural vibrations are similar to those obtained when an external load is applied at the robot end-effector only, each link of the structure will deform due to the stresses transmitted through the robot joints at points $O_j$. As a result, the deformations $\boldsymbol{\varepsilon}_j$ of the beam cross-section can be approximated by the deformations of a tip-loaded beam

$$\boldsymbol{\varepsilon}_j = \mathrm{diag}(f_j, g_j, g_j, f_j, h_j, h_j)\, \delta\boldsymbol{\theta}_j \tag{20}$$

where $\delta\boldsymbol{\theta}_j = \boldsymbol{\varepsilon}_j(x = L_j)$ represents the deformation of the beam at its tip and

$$f_j(x) = x/L_j,\ g_j(x) = 0.5x^2(3L_j - x)/L_j^3,\ h_j(x) = 2x(L_j - 0.5x)/L_j^2 \tag{21}$$

As a result, the global displacement $\boldsymbol{\delta}_j$ of the beam cross-section at $x$ can be expressed as a sum of two terms:

$$\boldsymbol{\delta}_j = \begin{bmatrix} \mathbf{I}_3 & \mathbf{D}_{(\times)} \\ \mathbf{0}_3 & \mathbf{I}_3 \end{bmatrix} \delta\mathbf{t}_{j-1} + \boldsymbol{\varepsilon}_j,\ \text{with } D_{(\times)} = \begin{bmatrix} 0 & 0 & 0 \\ 0 & 0 & -x \\ 0 & x & 0 \end{bmatrix} \tag{22}$$

In this sum, the left terms corresponds to the displacement of the undeformed beam due to the displacement of the node located at $O_j$.

Introducing (20) to (22) into (9) leads to the following equation:

$$T_{jl} = 1/2\, \omega_l^2 \cos^2(\omega_l t + \varphi_l) \left( \begin{bmatrix} \delta\mathbf{t}_{j-1}^T & \delta\mathbf{t}_j^T \end{bmatrix} \mathbf{M}_j^{red} \begin{bmatrix} \delta\mathbf{t}_{j-1} \\ \delta\mathbf{t}_j \end{bmatrix} \right) \tag{23}$$

where the expressions of each components of matrix $\mathbf{M}_j^{red}$ are given in [10].

Using these results, the total kinetic energy of the system for the $l$-th node is:

$$T_l = \sum_j T_{jl} = 1/2\, \omega_l^2 \cos^2(\omega_l t + \varphi_l)\, \delta\mathbf{t}^T \mathbf{M}_{tot}\, \delta\mathbf{t} \tag{24}$$

with $\mathbf{M}_{tot} = \mathrm{diag}(\mathbf{M}_1^{red}, \ldots, \mathbf{M}_n^{red})$ and $\delta\mathbf{t}^T = [\delta\mathbf{t}_0^T, \delta\mathbf{t}_1^T, \ldots, \delta\mathbf{t}_{n-1}^T, \delta\mathbf{t}_n^T]$

Then, assuming that the first natural modes of vibrations, i.e. the modes that have the most energy, lead to deformations that are close to the static deformations of the robot under a load applied on the end-effector, the mass matrix can be recomputed into the Cartesian coordinates associated with the tool end-point using the Jacobian matrix $\mathbf{J}_\theta$ defined at expression (3) (which depend on the robot configuration $\mathbf{q}$ and computed with respect to virtual joint coordinates $\boldsymbol{\theta}$) using following expression

$$\mathbf{M}_C = \mathbf{J}_\theta^T \mathbf{M}_\theta \mathbf{J}_\theta \qquad (25)$$

Thus, using expressions (25), it is possible to compute the reduced mass matrix $\mathbf{M}_C$ for a given robot configuration $\mathbf{q}$. The performances of this model reduction are shown in [10].

## 4      Machining process

Let us obtain the model of the cutting force which depends on the relative position of the tool with respect to the workpiece at each instant of machining. As follows from previous works [12], for the known chip thickness $h$, the cutting force $F_c$ can be expresses as

$$F_c(h) = k_0 \frac{h/h_s + r(h/h_s)^2}{1 + h/h_s} a_p, \quad h \geq 0 \qquad (26)$$

where $a_p$ is a depth of cut, $r = k_\infty/k_0 < 1$ depends on the parameters $k_\infty$, $k_0$ that define the so called stiffness of the cutting process for large and small chip thickness $h$ respectively (Fig. 3) and $h_s$ is a specific chip thickness, which depends on the current state of the tool cutting edge. The parameters $k_0$, $h_s$, $r$ are evaluated experimentally for a given combination of tool/working material. To take into account the possible loss of contact between the tool and the workpiece, the above expression should be supplement by the case of $h < 0$ as

$$F_c(h) = 0, \quad \text{if} \quad h < 0 \qquad (27)$$

For the multi-edge tool the machining surface is formed by means of several edges simultaneously. The number of working edges varies during machining and depends on the width of cut. For this reason, the total force $F_c$ of such interaction is a superposition of forces $F_{c,i}$ generated by each tool edge $i$, which are currently in the contact with the workpiece. Besides, the contact force $F_{c,i}$ can be decomposed by its *radial* $F_{r,i}$ and *tangential* $F_{t,i}$ components (Fig. 4). In accordance with Merchant's model [13], the *t*-component of cutting force $F_{t,i}$ can be computed with the equation (26). The *r*-component $F_{r,i}$ is related with $F_{t,i}$ by following expression [14]

$$F_{r,i} = k_r F_{t,i} \qquad (28)$$

where the ratio factor $k_r$ depends on the given tool/workpiece characteristics.

It should be mentioned that in robotic machining it is more suitable to operate with forces expressed in the robot tool frame $\{x,y,z\}$. Then, the corresponding components $F_x$, $F_y$ (Fig. 4) of the cutting force $F_c$ can be expressed as follows

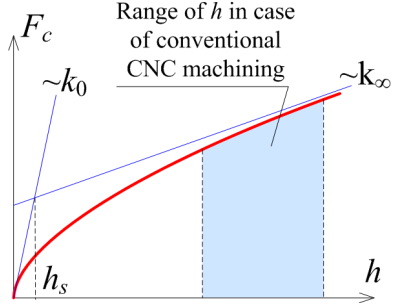
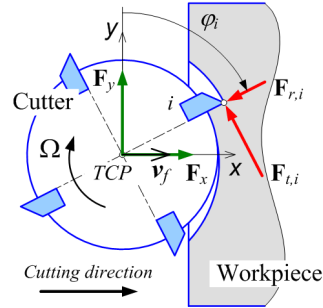

**Fig. 3.** Fractional cutting force model $F_c(h)$

**Fig. 4.** Forces of tool/workpiece interaction

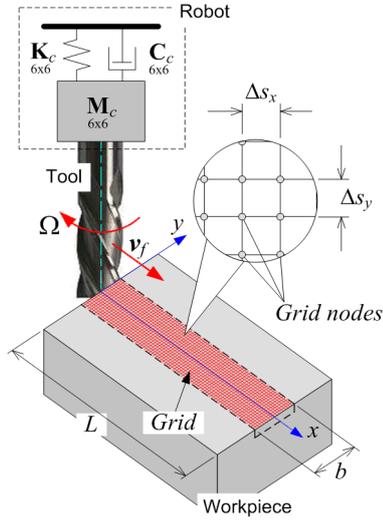
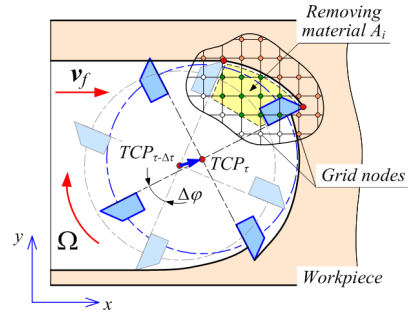

**Fig. 5.** Meshing of the workpiece area

**Fig. 6.** Evaluating the tool/workpiece intersection $A_i$ and computing the corresponding chip thickness $h_i$

$$F_x = \sum_{i=1}^{n_z} -F_{r,i} \cos \varphi_i + \sum_{i=1}^{n_z} F_{t,i} \sin \varphi_i$$
$$F_y = \sum_{i=1}^{n_z} F_{r,i} \sin \varphi_i + \sum_{i=1}^{n_z} F_{t,i} \cos \varphi_i \tag{29}$$

where $n_z$ is the number of currently working cutting edges, $\varphi_i$ is the angular position of the *i*-th cutting edge (the cutting force in *z* direction $F_z$ is negligible here). So, the vector of external loading of the robot due to the machining process can be composed in the frame {*x*,*y*,*z*} using the defined components $F_x$, $F_y$ as $\mathbf{F}=[F_x,F_y,0,0,0,0]^T$.

It should be stressed that the cutting force components $F_{r,i}$, $F_{t,i}$ mentioned in equation (26), (28) are computed for the given chip thickness $h_i$, which should be also evaluated. Let us define model for $h_i$ using mechanical approach. Then the chip thickness $h_i$ removed by $i$-th tooth depends on the angular position $\varphi_i$ of this tooth and it can be evaluated using to the geometrical distance between the position of the given tooth $i$ and the current machining profile (Fig. 4). It should be mentioned, that the main issue here is to follow the current relative position between the $i$-th tooth and the working material or to define whether the $i$-th tooth is involved in cutting for given instant of process. Because of the robot dynamic behavior and the regenerative mechanism of surface formation [15] this problem cannot be solved directly using kinematic relations. In this case it is reasonable to introduce a special rectangular grid, which decomposes the workpiece area into segments and allows tracking the tool/workpiece interaction and the formation of the machining profile (Fig. 5).

Here, Steps $\Delta s_x$, $\Delta s_y$ between grid nodes are constant and depend on the tool geometry, cutting condition and time discretization $\Delta \tau$. Each node $j$ ( $j = 1, N_w$ , $N_w$ is the number of nodes) of the grid can be marked as "1" or "0": "1" corresponds to nodes situated in the workpiece area with material (rose nodes in Fig. 6), "0" corresponds to nodes situated in workpiece area that was cut away (white nodes in Fig. 6).

In order to define the number of currently cut nodes by the $i$-th tooth, the previous instant of machining process should be considered. Let us define $A_i$ as an amount of working material that is currently cut away by the $i$-th tooth (Fig. 6). So, if node $j$ marked as "1" is located inside the marked sector (green nodes in Fig. 6), it changes to "0" and $A_i$ is increasing by $\Delta s_x \Delta s_y$. Analyzing all potential nodes and computing $A_i$, the chip thickness $h_i$, removed at given instant of the process by the *i*-th tooth, can be estimated by $h_i = A_i / R \Delta \alpha_i$, $i = 1, N_z$. The angle $\Delta \varphi_i$ determines the current angular position of the *i*-th tooth regarding to its position at the instant $\tau$-$\Delta\tau$ and referred to the position of TCP at $\tau$-$\Delta\tau$.

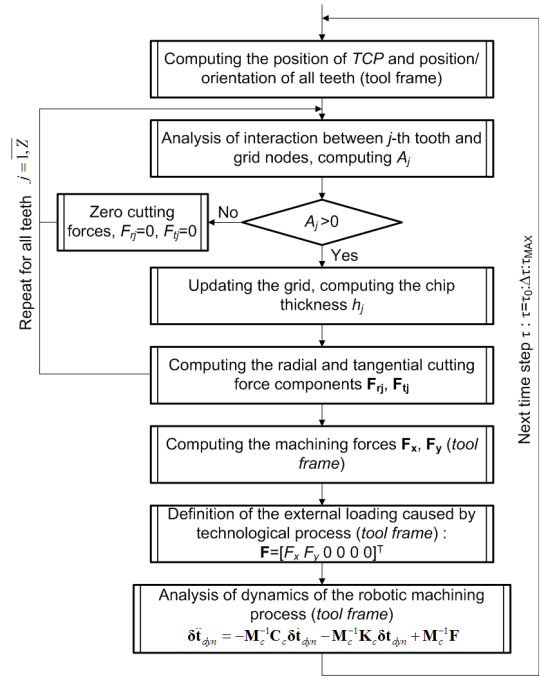

**Fig. 7.** Algorithm for numerical simulation of robotic machining process dynamics

Described mechanism of chip formation and the machining force model (26) allow computing the dynamic behavior of the robotic machining process where models of robot inertia and stiffness are discussed in the section 3 of the paper. The detailed algorithm that is used in numerical analysis is presented in Fig. 7, where the analysis of the robot dynamics is performed in the tool frame with respect to the dynamic displacement of the tool $\delta t_{dyn}$ fixed on the robot end-effector around its position on the trajectory.

## 5 Compliance error compensation technique

In industrial robotic controllers, the manipulator motions are usually generated using the inverse kinematic model that allows us to compute the input signals for actuators $\rho_0$ corresponding to the desired end-effector location $t_0$, which is assigned assuming that the compliance errors are negligible. However, if the external loading $F$ is essential, the kinematic control becomes non-applicable because of changes in the end-effector location. It can be computed from the non-linear compliance model as

$$\mathbf{t}_F = f^{-1}(\mathbf{F} \mid \mathbf{t}_0) \qquad (30)$$

where the subscripts 'F' and '0' refer to the loaded and unloaded modes respectively, and '|' separates arguments and parameters of the function $f(\ )$. Some details concerning this function are given in our previous publication [7].

To compensate this undeterred end-effector displacement from $\mathbf{t}_0$ to $\mathbf{t}_F$, the target point should be modified in such a way that, under the loading $\mathbf{F}$, the end-platform is located in the desired point $\mathbf{t}_0$. This requirement can be expressed using the stiffness model in the following way

$$\mathbf{F} = f\left(\mathbf{t}_0 \mid \mathbf{t}_0^{(F)}\right) \qquad (31)$$

where $\mathbf{t}_0^{(F)}$ denotes the modified target location. Hence, the problem is reduced to the solution of the nonlinear equation (31) for $\mathbf{t}_0^{(F)}$, while $\mathbf{F}$ and $\mathbf{t}_0$ are assumed to be given. It is worth mentioning that this equation completely differs from the equation $\mathbf{F} = f(\mathbf{t} \mid \mathbf{t}_0)$, where the unknown variable is $\mathbf{t}$. It means that here the compliance model does not allow us to compute the modified target point $\mathbf{t}_0^{(F)}$ straightforwardly, while the linear compensation technique directly operates with Cartesian compliance matrix [16].

To solve equation (31) for $\mathbf{t}_0^{(F)}$, similar numerical technique can be applied. It yields the following iterative scheme

$$\mathbf{t}_0^{(F)\prime} = \mathbf{t}_0^{(F)} + \alpha \cdot \left(\mathbf{t}_0 - f^{-1}(\mathbf{F} \mid \mathbf{t}_0^{(F)})\right) \qquad (32)$$

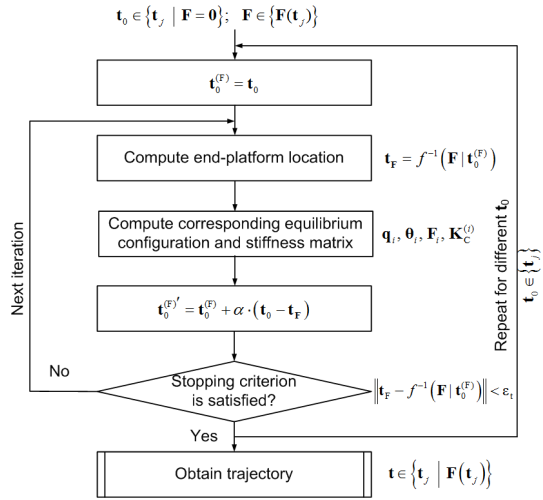

**Fig. 8.** Procedure for compensation of compliance errors

*(a) Off-line: modification of the target trajectory*

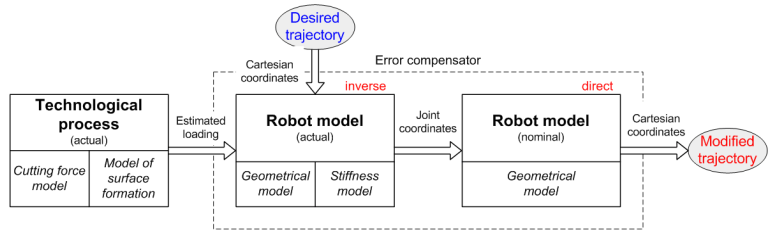

*(b) On-line: machining using revised trajectory*

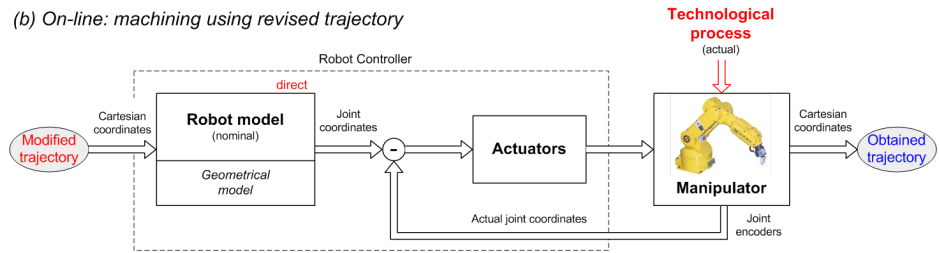

**Fig. 9.** Implementation of compliance error compensation technique

where the prime corresponds to the next iteration, $\alpha \in (0,1)$ is the scalar parameter ensuring the convergence. More detailed presentation of the developed iterative routines is given in Fig. 8.

Hence, using the proposed computational techniques, it is possible to compensate a main part compliance errors by proper adjusting the reference trajectory that is used as an input for robotic controller. In this case, the control is based on the inverse kinetostatic model (instead of kinematic one) that takes into account both the manipulator geometry and elastic properties of its links and joints. Implementation of developed compliance error compensation technique presented in Fig. 9.

## 6 Experimental verification

The developed compliance error compensation technique has been verified experimentally for robotic milling with the KUKA KR270 robot along a simple trajectory in aluminum workpiece. It is assumed that at the beginning of the technological process the robot is in the configuration $\mathbf{q}$ (see Table 1, Fig. 10). The parameters of the stiffness model for the considered robot have been identified in [17] and are presented in Table 1. Link masses required for the mass matrix of the robot are presented also in Table 1.

For the milling, the cutter with the external diameter $D=20$ mm and four teeth ($N_z=4$) distributed uniformly over the tool is used. For the given combination of the tool and the workpiece material the following parameters correspond to the cutting force model defined in (26): $k_0=5\times10^6$ N/m, $h_s=1.8\times10^{-5}$ m, $r=0.1$, $k_r=0.3$.

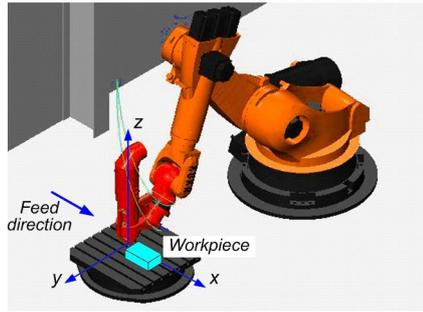 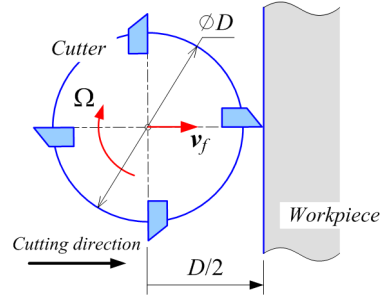

**Fig. 10.** Starting pose of the KUKA KR270 robot to perform the operation of milling

**Fig. 11.** Starting relative position of the tool with respect to the workpiece

**Table 1.** Initial data for robotic-based milling

| Joint coordinates, [deg] | | | | | |
|---|---|---|---|---|---|
| $q_1$ | $q_2$ | $q_3$ | $q_4$ | $q_5$ | $q_6$ |
| 90 | -50 | 120 | 180 | 25 | 180 |
| Joint compliances, [rad/N m]*10⁻⁶ | | | | | |
| $k_1$ | $k_2$ | $k_3$ | $k_4$ | $k_5$ | $k_6$ |
| 0.26 | 0.15 | 0.26 | 1.79 | 1.52 | 2.13 |
| Link masses, [kg] | | | | | |
| $m_1$ | $m_2$ | $m_3$ | $m_4$ | $m_5$ | $m_6$ |
| 336.8 | 259.4 | 85.2 | 54.5 | 36.3 | 18.2 |

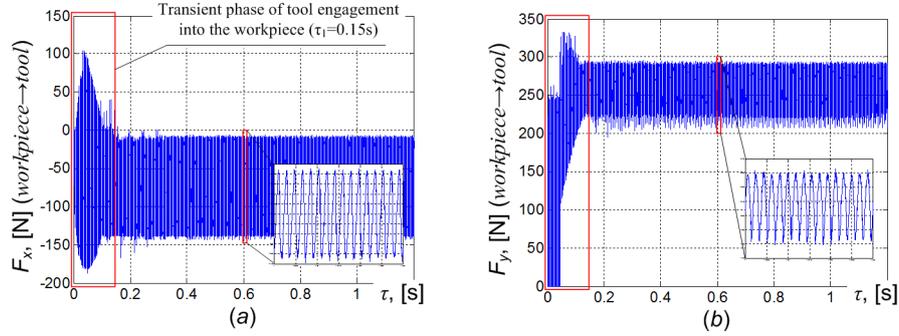

**Fig. 12.** Variation of machining force components $F_x$ (a) and $F_y$ (b) for whole milling process

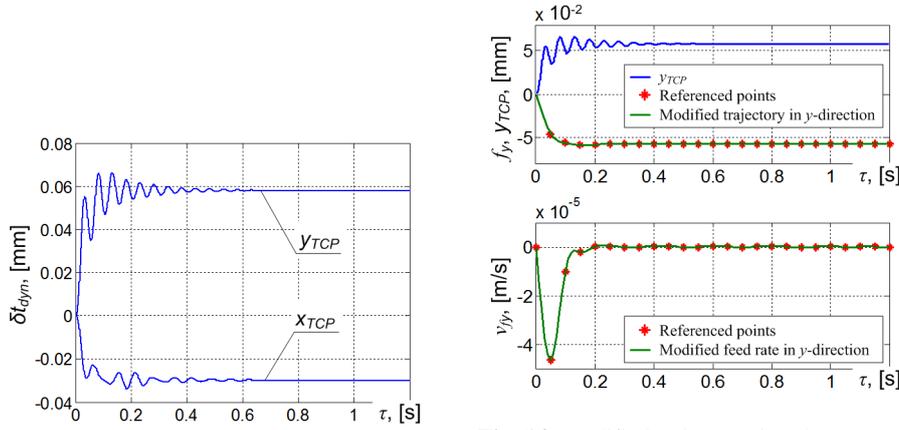

**Fig. 13.** Evolution of the tool dynamic displacement $\delta t_{dyn}$ that is composed from $x_{TCP}$ and $y_{TCP}$ components

**Fig. 14.** Modified trajectory $f_y$ and corresponding feed rate $v_{fy}$ in $y$-direction, computed based on the original dynamic displacement of the tool $\delta t_{dyn}$

Taking into account that the workpiece has a straight borders let us assume that at the instant t=0 one of the teeth of the tool is in contact with the workpiece material as it is shown in the Fig. 11. It is also assumed that the machining process is performing with the constant feed rate $v_f$=4 m/min (applied in $x$-direction of the robot tool frame) and the constant spindle rotation $\Omega$=8000 rpm along the straight line of 80 mm. Experimental verification and numerical simulation of the described case of the milling process with KUKA KR-270 robot using the algorithm shown in Fig. 7 allows us to trace the evolution of machining force x,y-components for the whole process (Fig. 12). The corresponding dynamic displacement of the tool around its current position on the trajectory is shown in Fig. 13.

In accordance with the obtained results the system robot/machining process realize complex vibratory motion. The high frequency component of this motion (about 700 Hz, Fig. 12) is related to the spindle rotation and the number of tool teeth $N_z$. In cer-tain cases such behavior can excites the dynamics of the robot (natural modes) but this study remains out the frame of the presented paper. On the contrary, the low frequency component of robot/tool motion (about 7 Hz, Fig. 13), especially in the *y*-direction (that is perpendicular to the applied feed) influences directly the quality of final product. Such motion is related to the robot compliance and it can be compensated using the error compensation technique described in the paper. Hence, let us form the modified trajectory based on the dynamic displacement of the robot end-effector in the *y*-direction (Fig. 14).

It should be stressed that the time step between referenced points of this modified trajectory is limited with the characteristics of the controller used in the robot (in the presented case this step is chosen 0.05 sec). The corresponding feed rate $v_{fy}$ for the modified trajectory has been computed. So, this new data (feed $f_y$ and feed rate $v_{fy}$) with the data defined in the beginning of this section allow us to compensate the trajectory error during machining caused by the robot compliance. The resulted compensated trajectory in the y-direction (in time domain) is presented in Fig. 15.

It should be noted that the part of the trajectory while machining tool is engaging into the workpiece does not have effect on the quality of final product (surface). During this stage the contact area between the tool and the workpiece is increasing progressively. Hence, at each instant of processing the cutter corrects the machining profile and eliminates trajectory errors produced during all previous instants. On the contrary, during the stage of machining with the fully engaged tool the trajectory in x,y-directions define directly the final machining profile and this part of trajectory is analyzed here (Fig. 15). Comparison results presented in Fig. 13 and Fig. 15 are summarized in Table 2. So after applying error compensation technique the static deviation in y direction has been reduced from 0.058 mm to 0.00014 mm (99.8%). Maximum defilation in the machining profile has been reduced from 0.063 mm to 0.0047 mm (92.6%). Low frequency remained the same for both cases.

Hence, obtained results show that the developed compliance error compensation allows us significantly increase the accuracy of the robotic-based machining.

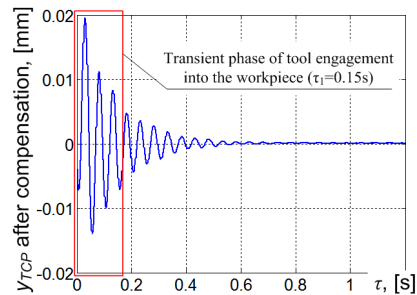

**Fig. 15.** Evolution of the dynamic displacement obtained after involving the error compensation technique into the analysis of robotic milling process

Table 2. Milling trajectory accuracy before and after compliance error compensation

| Performance measure | Original trajectory | Modified trajectory |
|---|---|---|
| Low frequency, [Hz] | 6.70 | 6.70 |
| Static deviation $y_s$, [mm] | 58.1e-3 | 0.14e-3 |
| Max deviation $y_{MAX}$, [mm] | 63.2e-3 | 4.70e-3 |

## 7 Conclusion

In robotic-based machining, an interaction between the workpiece and technological tool causes essential deflections that significantly decrease the manufacturing accuracy. Relevant compliance errors highly depend on the manipulator configuration and essentially differ throughout the workspace. Their influence is especially important for heavy serial robots. To overcome this difficulty this paper presents a new technique for compensation of the compliance errors caused by technological process. In contrast to previous works, this technique is based on the non-linear stiffness model and the reduced elasto-dynamic model of the robotic based milling process.

The advantages and practical significance of the proposed approach are illustrated by milling with of KUKA KR270. It is shown that after error compensation technique significantly increase the accuracy of milling. In future the proposed technique will be integrated in a software toolbox.

## 8 Acknowledgements

The authors would like to acknowledge the financial support of the ANR, France (Project ANR-2010-SEGI-003-02-COROUSSO) and the Region "Pays de la Loire", France.